\documentclass[11pt,a4paper]{article}

\usepackage[hyperref]{acl2017}
\usepackage{times}
\usepackage{latexsym}
\usepackage{CJKutf8}
\usepackage{url}
\usepackage{array}
\usepackage{graphicx}
\usepackage{multicol}
\usepackage{multirow}
\usepackage{float}

\aclfinalcopy 
\def\aclpaperid{347} 

\title{Flexible and Creative Chinese Poetry Generation Using Neural Memory}

\author{
Jiyuan Zhang$^{1,2}$, Yang Feng$^{3,1,4}$, Dong Wang$^{1}$\footnotemark[1], \\
\textbf{Yang Wang$^{1}$, Andrew Abel$^{6}$, Shiyue Zhang$^{1,5}$, Andi Zhang$^{1,5}$}\\
  $^1$Center for Speech and Language Technologies(CSLT), RIIT, Tsinghua University, China\\
  $^2$Shool of Software \& Microelectronics, Peking University, China \\
  $^3$Key Laboratory of Intelligent Information Processing,Institute of Computing Technology,CAS \\
  $^4$Huilan Limited, Beijing, China  \\
  $^5$Beijing University of Posts and Telecommunications, China \\
  $^6$Xi'an Jiaotong-Liverpool University, China\\
   {\tt zhangjy\_ml@pku.edu.cn},
   {\tt wangdong99@mails.tsinghua.edu.cn}
  \\
  }

\date{}

\begin{document}

\maketitle
\footnotetext[1]{Corresponding author: Dong Wang; RM 1-303, FIT BLDG, Tsinghua University, Beijing (100084), P.R. China.}

\begin{abstract}

It has been shown that Chinese poems can be successfully generated by \mbox{sequence-to-sequence} neural models, particularly with the attention mechanism.  A potential problem of this approach, however, is that neural models can only learn abstract rules, while poem generation is a highly creative process that involves not only rules but also innovations for which pure statistical models are not appropriate in principle.  This work proposes a memory-augmented neural model for Chinese poem generation, where the neural model and the augmented memory work together to balance the requirements of linguistic accordance and aesthetic innovation, leading to innovative generations that are still rule-compliant. In addition, it is found that the memory mechanism provides interesting flexibility that can be used to generate poems with different styles.

\end{abstract}

\section{Introduction}

Classical Chinese poetry is a special cultural heritage with over 2,000 years of history and is still fascinating us today.  Among the various genres, perhaps the most popular one is the quatrain, a special style with a strict structure (four lines with five or seven characters per line), a regulated rhythmical form (the last characters in the second and fourth lines must follow the same rhythm), and a required tonal pattern (tones of characters in some positions should satisfy a predefined regulation)~\cite{Wang:02}. This genre flourished mostly in the Tang Dynasty, and so are often called `Tang poems'. An example of a quatrain written by Wei Wang, a famous poet in the Tang Dynasty, is shown in Table~\ref{tang-quatrain}.

Due to the stringent restrictions in both rhythm and tone, it is not trivial to create a fully \mbox{rule-compliant} quatrain. More importantly, besides such strict regulations, a good quatrain should also read fluently, hold a consistent theme, and express a unique affection.
Therefore, poem generation is widely recognized as a very intelligent activity and can be performed only by knowledgeable people
with a lot of training.

\begin{table}[h]
\begin{center}
\begin{tabular}{|p{6cm}<{\centering}|}
\hline
\begin{CJK*}{UTF8}{gbsn}
乐游原 
\end{CJK*}\\
Climbing the Paradise Mound \\
\begin{CJK*}{UTF8}{gbsn}
向晚意不适，
\end{CJK*}
(* Z Z P Z) \\
As I was not in a good mood this evening round, \\
\begin{CJK*}{UTF8}{gbsn}
驱车登古原。
\end{CJK*}
(P P P Z P) \\
I went by cart to climb the Ancient Paradise Mound. \\
\begin{CJK*}{UTF8}{gbsn}
夕阳无限好，
\end{CJK*}
(* P P Z Z) \\
It is now nearing dusk, \\
\begin{CJK*}{UTF8}{gbsn}
只是近黄昏。
\end{CJK*}
(* Z Z P P) \\
When the setting sun is infinitely fine, which is a must. \\
\hline
\end{tabular}
\end{center}
\caption{\label{tang-quatrain} An example of a 5-char quatrain. The tonal pattern is shown at the end of each line, where 'P' indicates a level tone, 'Z' indicates a downward tone, and '*' indicates the tone can be either. The translation is from~\cite{tangyihe:05}.}
\end{table}

In this paper we are interested in machine poetry generation. Several approaches have been studied by researchers. For example, rule-based methods~\cite{Zhou:10}, statistical machine translation (SMT) models~\cite{Jiang:08,He:12} and neural models~\cite{Zhang:14,Wang-bics:16,Wang:planning}. Compared to previous approaches (e.g., rule-based or SMT), the neural model approach tends to generate more fluent poems and some generations are so natural that even professional poets can not tell they are the work of machines~\cite{Wang-bics:16}.

In spite of these promising results, neural models suffer from a particular problem in poem
generation, a lack of innovation. Due to the statistical nature of neural models, they pay much more attention to high-frequency patterns, whereas they ignore low-frequency ones.
In other words, the more regular and common the patterns, the better the neural model is good at learning them and tends to use them more frequently at run-time.
This property certainly helps to generate fluent sentences, but it is not always useful: the major value of poetry is not fluency, but the aesthetic innovation that can stimulate some unique feelings.
This is particularly true for Chinese quatrains that are highly compact and expressive: it is nearly impossible to find two
similar works in the thousands of years of history in this genre, demonstrating the importance of uniqueness or innovation.
Ironically, the most important thing, innovation, is largely treated as trivial, if not noise, by present neural models.

Actually this problem is shared by all generation models based on statistics (although it is more serious for neural models) and has aroused  a \mbox{long-standing} criticism for machine poem generation: it can generate, and sometimes generate well, but the generation tends to be unsurprising and not particularly interesting.
More seriously, this problem exists not only in poem generation, but also in all generation tasks that require innovation.

This paper tries to solve this extremely challenging problem. We argue that the essential problem is that statistical models are good at learning general rules (usage of regular words and their combinations) but are less capable of remembering special instances that are difficult to cover with general rules. In other words, there is only rule-based reasoning, no instance-based memory. We therefore present a memory-augmented neural model which involves a neural memory so that special instances can be saved and referred to at run-time. This is like a human poet who creates poems by not only referring to common rules and patterns, but also recalls poems that he has read before. It is hard to say whether this combination of rules and instances produces true innovation (which often requires real-life motivation rather than simple word reordering), but it indeed offers interesting flexibility to generate new outputs that look creative and are still rule-compliant. Moreover, this flexibility can be used in other ways, e.g., generating poems with different styles.

In this paper, we use the memory-augmented neural model to generate flexible and creative Chinese poems.  We investigate three scenarios where adding a memory may contribute: the first scenario involves a well trained neural model where we aim to promote innovation by adding a memory, the second scenario involves an over-fitted neural model where we hope the memory can regularize the innovation, and in the third scenario, the memory is used to encourage generation of poems of different styles.

\section{Related Work}

A multitude of methods have been proposed for automatic poem generation. The first approach is based on rules and/or templates. For example, phrase search~\cite{Tosa:09,Wu:09}, word association norm~\cite{Netzer:09}, template search~\cite{Oliveira:09}, genetic search \cite{Zhou:10}, text summarization~\cite{Yan:13}. Another approach involves various SMT methods, e.g., \cite{Jiang:08,He:12}.
A disadvantage shared by the above methods is that they are based on the surface forms of words or characters, having no deep understanding of the meaning of a poem.

More recently, neural models have been the subject of much attention. A clear advantage of the neural-based methods is that they can `discover' the meaning of words or characters, and can therefore more deeply understand the meaning of a poem. Here we only review studies on Chinese poetry generation that are mostly related to our research. The first study we have found in this direction is the work by~\citeauthor{Zhang:14}~\shortcite{Zhang:14}, which proposed an RNN-based approach that produces each new line \mbox{character-by-character} using a recurrent neural network (RNN), with all the lines generated already (in the form of a vector) as a contextual input. This model can generate quatrains of reasonable quality. \citeauthor{Wang:16}~\shortcite{Wang:16} proposed a much simpler neural model that treats a poem as an entire character sequence, and poem generation is conducted character-by-character. This approach can be easily extended to various genres such as Song \mbox{Iambics}. To avoid theme drift caused by this \mbox{long-sequence} generation, \citeauthor{Wang:16}~\shortcite{Wang:16} utilized the neural attention mechanism~\cite{bahdanau2014neural} by which human intention is encoded by an RNN to guide the generation.  The same model was used by~\citeauthor{Wang-bics:16}~\shortcite{Wang-bics:16} for Chinese quatrain generation. \citeauthor{yan2016poet}~\shortcite{yan2016poet} proposed a hierarchical RNN model that conducts iterative generation. Recently, \citeauthor{Wang:planning}~\shortcite{Wang:planning} proposed a similar sequence generation model, but with the difference that attention is placed not only on the human input, but also on all the characters that have been generated so far. They also proposed a topic planning scheme to encourage a smooth and consistent theme.

All the neural models mentioned above try to generate fluent and meaningful poems, but none of them consider innovation. The \mbox{memory-augmented} neural model proposed in this study intends to address this issue. Our system was built following the model structure and training strategy proposed by~\citeauthor{Wang-bics:16}~\shortcite{Wang-bics:16} due to its simplicity and demonstrated quality, but the memory mechanism is general and can be applied to any of the models presented above.

The idea of memory argumentation was inspired by the recent advance in neural Turing machine~\cite{graves2014neural,graves2016hybrid} and memory network~\cite{weston2014memory}.
These new models equip neural networks with an external memory that can be accessed and manipulated via some \emph{trainable} operations.
In comparison, the memory in our work plays a simple role of knowledge storage, and the only operation is simple pre-defined READ.
In this sense, our model can be regarded as a simplified neural Turing machine that omits training.









\section{Memory-augmented neural model}
\label{sec:method}

In this section, we first present the idea of memory augmentation, and then describe the model structure and training method.

\subsection{Memory augmentation}

\begin{figure}
	\centering
	\includegraphics[width=3in]{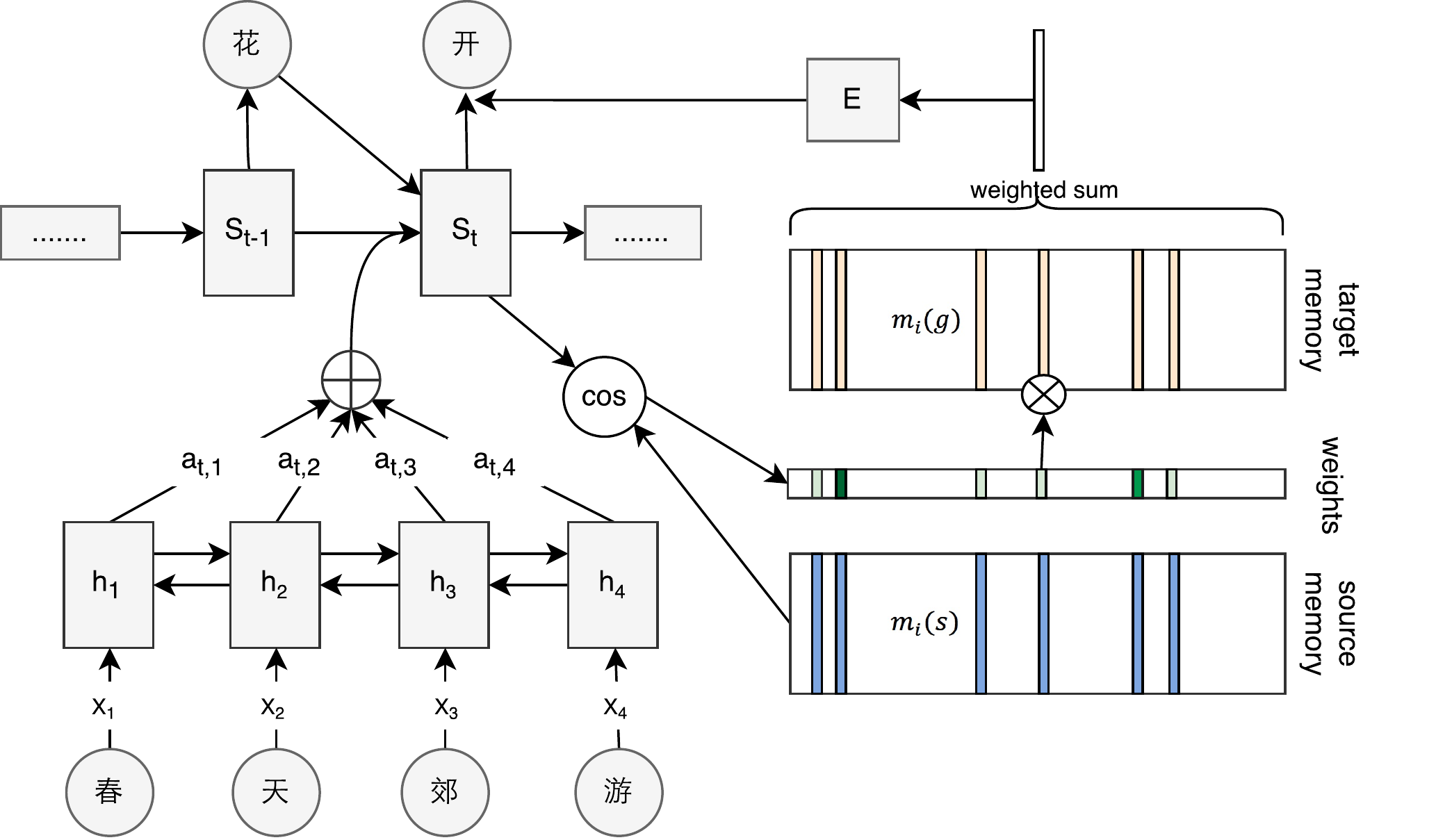}
\vspace{-4mm}
	\caption{The memory-augmented neural model used for Chinese poetry generation.}
	\label{fig:model}
\end{figure}

The idea of memory augmentation is illustrated in Fig.~\ref{fig:model}. It contains two components, the neural model component on the left, and the memory component on the right. In this work, the attention-based RNN generation model presented by~\cite{Wang-bics:16} is used as the neural model component, although any neural model is suitable. The memory component involves a set of `direct' mappings from input to output, and therefore can be used to memorize some special cases of the generation that can not be represented by the neural model. For poem generation, the memory stores the information regarding which character should be generated in a particular context. The output from the two components are then integrated, leading to a consolidated output.

There are several ways to understand the memory-augmented neural model. Firstly, it can be regarded as a way of combining reasoning (neural model) and knowledge (memory).  Secondly, it can be regarded as a way of combining \mbox{rule-based} inference (neural model) and instance-based retrieval (memory). Thirdly, it can be regarded as a way of combining predictions from complementary systems, where the neural model is continuous and parameter-shared, while the memory is discrete and contains no parameter sharing.  Finally, the memory can be regarded as an effective regularization that constrains and modifies the behavior of the neural model, resulting in generations with desired properties.
Note that this \mbox{memory-augmented} neural model is inspired by and related to the memory network proposed by~\citeauthor{weston2014memory}\shortcite{weston2014memory} and \citeauthor{graves2016hybrid}\shortcite{graves2016hybrid}, but we more focus on an accompanying
memory that plays the role of assistance and regularization.

\subsection{Model structure}

Using the Chinese poetry generation model shown in Fig.~\ref{fig:model} as an example, this section discusses the creation of a
memory-augmented neural model. Firstly, the neural model part is an attention-based sequence-to-sequence model~\cite{bahdanau2014neural}. The encoder is a bi-directional RNN (with GRU units) that converts the input topic words, denoted by the embeddings of the compositional characters $(x_{1}, x_{2}, ...,x_{N})$, into a sequence of hidden states $(h_{1} , h_{2} , ...,h_{N})$. The decoder then generates the whole quatrain \mbox{character-by-character}, denoted by the corresponding embeddings $(y_{1}, y_{2}, ...)$. At each step t, the prediction for the state $s_{t}$ is based on the last generation $y_{t-1}$, the previous status $s_{t-1}$ of the decoder, as well as all the hidden states $(h_{1} , h_{2} , ... )$ of the encoder. Each hidden state $h_{i}$ contributes to the generation according to a relevance factor $\alpha_{t}$ that measures the similarity between $s_{t-1}$ and $h_{i}$. This is written as:

\[
s_{t} = f_d (y_{t-1}, s_{t-1}, \sum_{i=1}^{N} \alpha_{t,i} h_i )
\]

\noindent where $\alpha_{t,i}$ represents the contribution of $h_i$ to the present generation, and can be implemented as any function. The output of the model is a posterior probability over the whole set of characters, written by
\[
z_t=\sigma(s_t W)
\]
where $W$ is the projection parameter.

The memory consists of a set of elements $\{m_i\}_{i=1}^K$, where $K$ is the size of the memory. Each element $m_i$ involves two parts, the source part $m_i(s)$, that encodes the context, i.e. when this element should be selected, and the target part $m_i(g)$, that encodes what should be output if this element is selected. In our study, the neural model is firstly trained, and then the memory is created by running $f_d$ (the decoder of the neural model). Specifically, for the $k$-th poem selected to be in the memory, the character sequence is input to the decoder one by one, with the contribution from the encoder set to zero. Denoting the starting position of this poem in the memory is $p_k$, the status of the decoder at the $j$-th step is used as the source part of the $(p_k + j)$-th element of the memory, and the embedding of the corresponding character, $x_{j}$, is set to be the target part. this is formally written as:

\begin{equation}
\label{eq:ms}
m_{i}(s) = f_d (x_{j-1}, s_{j-1}, 0)
\end{equation}

\noindent and
\[
m_{i}(g) = x_{j}
\]

\noindent where
\[
i = p_k + j.
\]

At run-time, the memory elements are selected according to their fit to the present decoder status $s_{t}$, and then the outputs of the selected elements are averaged as the output of the memory component. We choose cosine distance to measure the fitting degree, and have\footnote{In fact, we run a parallel decoder to provide $s_{t}$ in Eq.(\ref{eq:mem}). This decoder does not accept
input from the encoder and so is consistent
with the memory construction process as Eq.(\ref{eq:ms}).}:

\begin{equation}
\label{eq:mem}
v_t = \sum_{i=1}^{K} cos(s_{t},m_i(s)) m_i(g).
\end{equation}

The output of the neural model and the memory can be combined in various ways. Here, a simple linear combination before the softmax is used, i.e.,
\begin{equation}
\label{eq:sigma}
z_t=\sigma(s_t W + \beta v_t E)
\end{equation}
\noindent where $\beta$ is a pre-defined weighting factor, and $E$ contains word embeddings of all the characters. Although it is possible to train $\beta$ from the data, we found that the learned $\beta$  is not better than the manually-selected one. This is
probably because $\beta$ is a factor to trade-off the contribution from the model and the memory, and how to make the trade-off should be
a `prior knowledge' rather than a tunable parameter. In fact, if it is trained, than it will be immediately adapted to match the training
data, which will nullify our effort to encourage innovative generation.



\subsection{Model Training}

In our implementation, only the neural model component is required to be trained. The training algorithm follows the scheme defined in~\cite{Wang-bics:16}, where the cross entropy between the distributions over Chinese characters given by the decoder and the ground truth
is used as the objective function. The optimization uses the \mbox{SGD} algorithm together with AdaDelta to adjust the learning rate~\cite{zeiler2012adadelta}.

\section{Memory augmentation for Chinese poetry generation}

This section describes how the memory mechanism can be used to trade-off between the requirements for rule-compliant generation and aesthetic innovation, and how it can also be used to do more interesting things, for example style transfer.

\subsection{Memory for innovative generation}

In this section, we describe how the memory mechanism promotes innovation. Monitoring the training process for the attention-based model, we found that the cost on the training set will keep decreasing until approaching zero, but on the validation set, the degradation stops after only one iteration. This can be explained by the fact that Chinese quatrains are highly unique, so the common patterns can be fully learned in one iteration, resulting in overfitting with additional iterations. Due to the overfitting, we observe that with the one-iteration model, reasonable poems can be generated, and with the over-fitted model, the generated poems are meaningless, in that they do not resemble feasible character sequences.

\begin{figure*}
	\centering
	\includegraphics[width=5in]{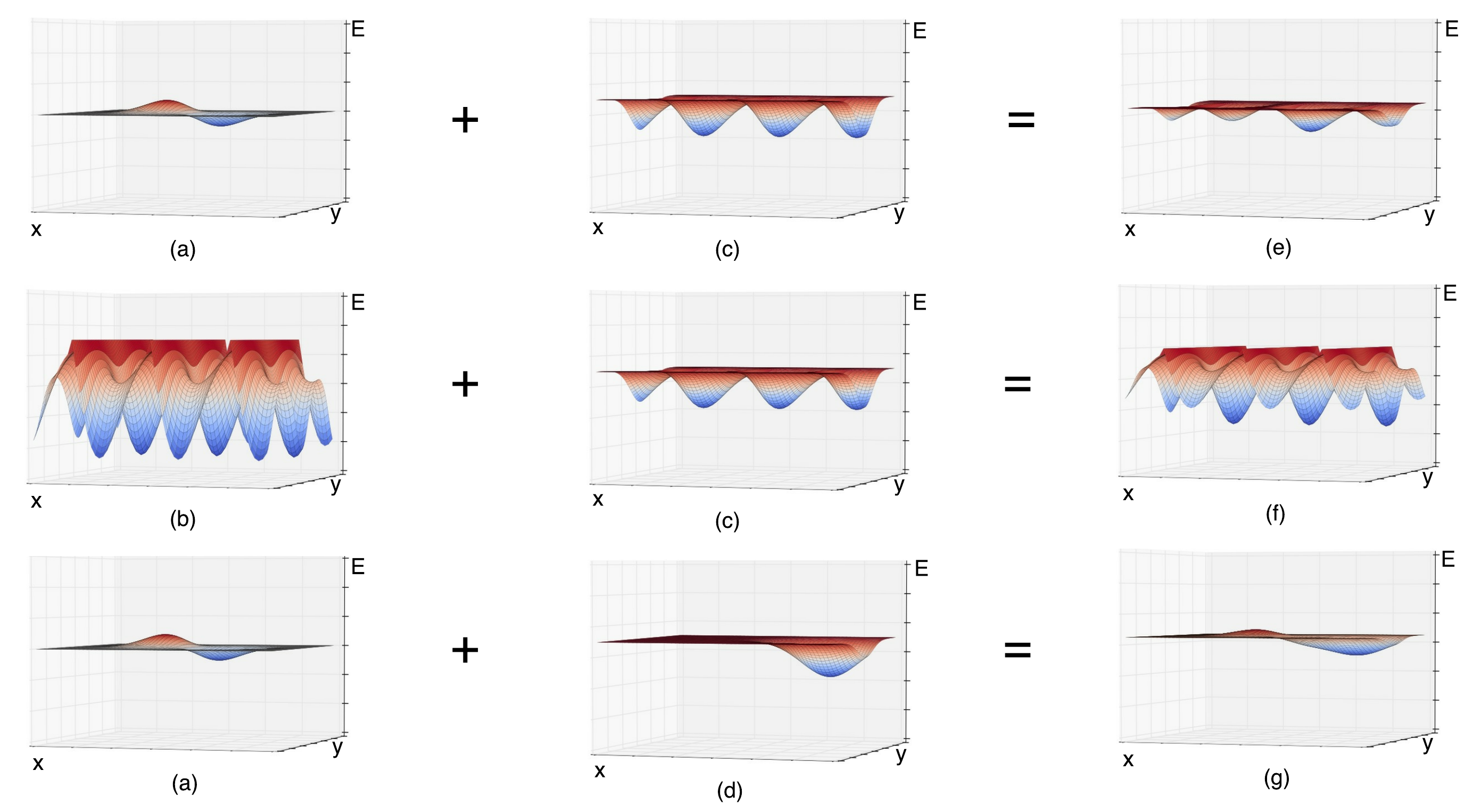}
	\caption{The energy surface for (a) one-iteration model (b) overfitted model (c) memory (d) style memory (e) one-iteration model augmented with memory (f) overfitted model augmented with memory (g) one-iteration model augmented with style memory.}
	\label{fig:theory-1}
\end{figure*}

The energy model perspective helps to explain this difference. For the one-iteration model, the energy surface is smooth and the energy of the training data is not very low, as illustrated in plot (a) in Fig.~\ref{fig:theory-1}, where the $x$-axis represents the input and $y$-axis represents the output, and the \mbox{$z$-axis} represents the energy. With this model, inputs with small variance will be attracted to the same \mbox{low-energy} area, leading to similar generations.  These generations are trivial, but at least reasonable.  If the model is overfitted, however, the energy at the locations of the training data becomes much lower than their surrounding areas, leading to a bumpy energy surface as shown in plot (b) in Fig.~\ref{fig:theory-1}. With this model, inputs with a small variation may be attracted to very different low-energy areas, leading to significantly different generations. Since many of the low-energy areas are nothing to do with good generations but are simply caused by the complex energy function, the generations can be highly surprising for human readers, and the quality is not guaranteed. In some sense, these generations can be regarded as `innovative' , but based on observations made in our experiments, most of them are meaningless.

The augmented memory introduces a new energy function, which is combined with the energy function of the neural model to change the energy surface of the generation system. This can be seen in Eq. (\ref{eq:sigma}),
where $s_tW$ and $\beta v_tE$ can be regarded as the energy function of the neural model component and the memory component, respectively, and the energy function of the memory-augmented system is the sum of the energy functions of these two components. For this reason, the effect of the memory mechanism can be regarded as a regularization of the neural model that will adjust its generation behavior.

This regularization effect is illustrated in Fig.~\ref{fig:theory-1}, where the energy function of the memory shown in plot (c)
is added to the energy function of the one-iteration model and the overfitted model, as shown in plot (e) and plot (f) respectively. It can be seen that with the memory involved, the energy surface becomes more bumpy with the \mbox{one-iteration} model, and more smooth with the overfitted model. In the former case, the effect of the memory is to encourage innovation, while still focusing on \mbox{rule-compliance}, and in the latter case, the effect is to encourage rule compliance, while keeping the capability for innovation.

It is important to notice that the energy function of the memory component is a linear combination of the energy
functions of the compositional elements (see Eq.(\ref{eq:mem})), each of which is convex and is minimized
at the location represented by the element. This means that the energy surface of the memory is rather `healthy',
in the sense that \mbox{low-energy} locations mostly correspond to good generations. For this
reason, the regularization provided by the memory is safe and helpful.









\subsection{Memory for style transfer}

The effect of the memory is easy to control. For example, the complexity of the behavior can be controlled by the memory size, the featured bias can be controlled by memory selection, and the strength of the impact can be controlled by the weighting parameter $\beta$.  This means that the memory mechanism is very flexible and can be used to produce poems with desired properties.

In this work, we use these capabilities to generate poems with different styles. This has been illustrated in Fig.~\ref{fig:theory-1}, where the energy function of the style memory shown in plot (d) is biased towards a particular style, and once it is added to energy function of the one-iteration model, the resulting energy function shown in plot (g) obtains lower values at locations corresponding to the
locations of the memory, which encourages generation of poems with similar styles as those poems in the memory.


\section{Experiments}

This section describes the experiments and results carried out in this paper. Here, The baseline system was a reproduction of the Attention-based system presented in ~\cite{Wang-bics:16}. the model in  This system has been shown to be rather flexible and powerful: it can generate different genres of Chinese poems, and when generating quatrains it has been shown to be able to fool human experts in many cases~\cite{Wang-bics:16} and the authors had did a thorough comparison with competitive methods mentioned in the related work of this paper. We obtained the database and the source code (in theano), and reproduced their system using Tensorflow from Google\footnote{https://www.tensorflow.org/}. We didn't make comparisons with some previous methods such as NNLM, SMT, RNNPG as they had
been fully compared in~\cite{Wang-bics:16} and all of them were much worse than the attention-based system. Another reason was that the experts were not happy to evaluate poems with clearly bad quality.
We also reproduced the model in~\cite{Wang:planning} with the help of the first author. However, since their implementation did not involve any restrictions on rhythm and tone, the experts were reluctant to recognize them as good poems.
With a larger dataset (e.g., $1$ Million poems), it is assumed that the rhythm and tone can be learned and their system would be good in both fluency and rule compliance. It should be also emphasized that the memory approach proposed in this paper is a general technique and is complementary to other efforts such as the planning approach~\cite{Wang:planning} and the recursive approach~\cite{yan2016poet}.

Based on the baseline system, we built the memory-augmented model, and conducted two experiments to
demonstrate its power. The first is an innovation experiment which employs memory to promote or regularize the generation of innovative poems, and the second is a style-transfer experiment which employs memory to generate flexible poems in different styles.

We invited $34$ experts to participate in the experiments, and all of them have rich experience not only evaluating poems, but also in writing them. Most of the experts are from prestigious institutes, including Peking university and the Chinese \mbox{Academy} of Social Science (CASS).  Following the suggestions of the experts, we use five metrics to evaluate the generation, as listed below:






\begin{itemize}
	\item Compliance: if regulations on tones and rhymes are satisfied;
	\item Fluency: if the sentences read fluently and convey reasonable meaning;
	\item Theme consistency: if the entire poem adheres to a single theme;
	\item Aesthetic innovation: if the quatrain stimulates any aesthetic feeling with elaborate innovation;
	\item Scenario consistency: if the scenario remains consistent.
\end{itemize}




\subsection{Datasets}

The baseline system was built with two customized datasets. The first dataset is a Chinese poem corpus (CPC), which we used in this work to train the embeddings of Chinese characters.  Our CPC dataset contains 284,899 traditional Chinese poems in various genres, including Tang quatrains, Song Iambics, Yuan Songs, and Ming and Qing poems. This large quantity of data ensures reliable learning for the semantic content of most Chinese characters.

Our second dataset is a Chinese quatrain corpus (CQC) that we have collected from the internet,
which consists of $13, 299$ 5-char quatrains and $65,560$ 7-char quatrains.  This corpus was used to train the attention-based RNN baseline.  We filtered out the poems whose characters are all \mbox{low-frequency} (less than $100$ counts in the database). After the filtering, the remaining corpus contains 9,195 5-char quatrains and 49,162 \mbox{7-char} quatrains. We used 9,000 5-char and 49,000 \mbox{7-char} quatrains to train the attention model, and the rest for validation.

Another two datasets were created for use in the memory-augmented system. Our first dataset, MEM-I, contains $500$ quatrains randomly selected from our CQC corpus. This dataset was used to produce the memory in the innovation experiment; the second dataset, MEM-S, contains $300$ quatrains with clear styles, including $100$ pastoral, $100$ battlefield and $100$ romantic quatrains. It was used to generate memory with different styles in the style-transfer experiment. All the datasets will be released online\footnote{http://vivi.cslt.org}.





\subsection{Evaluation Process}

We invited $34$ experts to evaluate the quality of the poem generation. In the innovation experiment, the evaluation consisted of a comparison between different systems and configurations in terms of the five metrics. The innovation questions presented the expert with two poems, and asked them to judge which of the poems was better in terms of the five metrics; in the style-transfer experiment, the evaluation was performed by identifying the style of a generated poem.  The evaluation was conducted online, with each questionnaire containing $11$ questions focusing on innovation and $4$ questions concerned with style-transfer. Each of the style-transfer questions presented the expert with a single poem and asked them to score it between $1$ to $5$, with a larger score being better, in terms of compliance, aesthetic innovation, scenario consistency, and fluency.  They were also asked to specify the style of the poem.

Using the poems generated by our systems, we generated many different questions of both types, and then created a number of online questionnaires that randomly selected from these questions.  This meant that as discussed above, each questionnaire had $11$ randomly selected innovation questions, and $4$ randomly selected style transfer questions.  Each question was only used once, meaning that it was not duplicated on multiple questionnaires, and so each questionnaire was different.

Experts could choose to answer multiple questionnaires if they wished, as each one was different.  From the $34$ experts, we collected $69$ completed questionnaires, which equals to $759$ innovation questions and $276$ style-transfer questions.


\subsection{Innovation experiment}

This experiment focuses on the contribution of memory for innovative poem generation. We experimented with two configurations: one is with a one-iteration model ($C_1$) and the other is with an overfitted model ($C_{\infty}$). The memory was generated from the $500$ quatrains in MEM-I, and the weighting factor was defined empirically as $16$ for $C_1$ and $49$ for $C_{\infty}$.

The topics of the generation were $160$ keywords randomly selected from Shixuhanyinge~\cite{liu1735book}. Given a pair of poems generated by two different configurations using the same topic, the experts were asked to choose which one they preferred. The evaluation is therefore pair-wised, and each pair of configurations contains at least $180$ evaluations. The results are shown in Table~\ref{tab:innov}, where the preference ratio for each pair of configurations was tested in terms of the $5$ metrics.

\begin{table*}[htb]
	
	\centering
	\fontsize{6.5}{8}\selectfont
	\begin{tabular}{|l|c|c|c|c|c|}
		\hline
		&\multicolumn{5}{c|}{Preference Ratio} \\
		\hline
		&Compliance&Fluency&Theme       &Aesthetic     & Scenario\\
		&          &       &Consistency &Innovation    &Consistency\\
		\hline
		$C_1$ vs $C_{\infty}$             &0.59:0.41&0.68:0.32 &0.70:0.30&0.68:0.32&0.69:0.31\\
		\hline
		$C_1$ vs $C_1$+Mem                &0.41:0.59&0.36:0.64&0.37:0.63&0.33:0.67&0.43:0.57\\
		$C_{\infty}$ vs $C_{\infty}$+Mem  &0.40:0.60&0.26:0.74&0.32:0.68&0.30:0.70&0.36:0.64\\
		\hline
		$C_1$ vs $C_{\infty}$+Mem         &0.43:0.57&0.58:0.42&0.59:0.41&0.50:0.50&0.59:0.41\\
		\hline
	\end{tabular}
	\caption{Preference ratios for systems with or without overfitting and with or without memory augmentation.}
	\label{tab:innov}
\end{table*}

From the first row of Table~\ref{tab:innov}, we observe that the experts have a clear preference for the poems generated by the $C_1$ model, the one that can produce fluent yet uninteresting poems.  In particular, the `aesthetic innovation' score for $C_{\infty}$ is not better than $C_1$, which was different from what we expected.  Informal offline discussions with the poetry experts found that the experts identified some innovative expression in the $C_{\infty}$ condition, but most of the them was regarded as being nonsense in the opinion of many of the experts. In comparison to sparking innovation, fluency and being meaningful is more important not only for \mbox{non-expert} readers, but also for professional poets.
In other words, only meaningful innovation is regarded as innovation, and irrational innovation is simply treated as junk.

From the second and third rows of Table~\ref{tab:innov}, it can be seen that involving memory significantly improves both $C_1$ and $C_{\infty}$, particularly for $C_{\infty}$. For $C_1$, the most substantial improvement is observed in terms of `Aesthetic innovation', which is consistent with our argument that memory can help encourage innovation for this model. For $C_{\infty}$, `Fluency' seems to be the most improved metric.  This is also consistent with our argument that involving memory constrains over-innovation for over-fitted models.

The last row of Table~\ref{tab:innov} is an extra experiment that investigates if $C_{\infty}$ is regularized well enough after introducing the memory. It seems that with the regularization, the overfitting problem is largely solved, and the generation is nearly as fluent and consistent as the $C_1$ condition.
Interestingly, the score for aesthetic innovation is also significantly improved. Since the regularization is not supposed to boost innovation, this seems confusing at first glance (in comparison to the result on the same metric in the first row), but this is probably because the increased fluency and consistency makes the innovation more appreciated, therefore doubly confirming our argument that true innovation should be reasonable and meaningful.









\subsection{Style-transfer experiment}

In the second experiment, the memory mechanism is used to generate poems in different styles. We chose three styles: pastoral, battlefield, and romantic. A style-specific memory, which we call style memory, was constructed for each style by the corresponding quatrains in the MEM-S dataset. The system with one-iteration model $C_1$ was used as the baseline. Two sets of topics were used in the experiment, one is general and the other is style-biased. The experiments then investigate if the memory mechanism can produce a clear style if the topic is general, and can transfer to a different style if the topic is style-biased already.
The experts were asked to specify the style from four options including the three defined above and a `unclear style' option.  In addition, the experts were asked to score the poems in terms of compliance, fluency, aesthetic innovation, and scenario consistency, which we can use to check if the style transfer impacts the quality of the poem generation.
Note that we did not ask for the theme consistency to be scored in this experiment because the topic words were not presented to the experts, in order to prevent the topic affecting their judgment regarding the style. The score ranges from $1$ to $5$, with a larger score being better.

Table~\ref{tab:style-general} presents the results with the general topics. The numbers show the probabilities that the poems generated by a particular system were labeled as having various styles. Since the topics are unbiased in types, the generation of the baseline system is assumed to be with unclear styles. For other systems, the style of the generation is assumed to be the same as the style of their memories. The results in Table~\ref{tab:style-general} clearly demonstrates these assumptions.
The tendency that romantic poems are recognized as pastoral poems is a little surprising. Further analysis shows that experts tend to recognize romantic poems as pastoral poems only if there are any related symbols such as trees, mountain, river. These words are very general in Chinese quatrains. The indicator words of romantic poems such as skirt, rouge, and singing are not as popular and their indication power is not as strong, leading to less labeling of romantic poems, as shown in the results.

\begin{table}[htb]
	\centering
	\fontsize{6.5}{8}\selectfont
	\begin{tabular}{|l|c|c|c|c|}
		\hline
		& \multicolumn{4}{c|}{Probability }       \\
		\hline
		Model                & Pastoral  & Battlefield & Romantic & Unclear \\
		\hline
		$C_1$ (Baseline)        &  0.09     &    0.04     &   0.18   &  0.69 \\
		$C_1$ + Pastoral Mem    &  0.94     &    0.00     &   0.06   &  0.00  \\
		$C_1$ + Battlefield Mem &  0.05     &    0.93     &   0.00   &  0.02  \\
		$C_1$ + Romantic Mem    &  0.17     &    0.00     &   0.61   &  0.22 \\
		\hline
	\end{tabular}
	\caption{\label{tab:style-general} Probability that poems generated
		by each configuration with general topics are labeled as various styles.}
\end{table}

We also tested transferring from one style to another. This was achieved by generating poems with some style-biased topics, and then using a style memory to force the generation to change the style. Our experiments show that in 73\% cases the style can be successfully transferred.



Finally, the scores of the poems generated with and without the style memories are shown in Table~\ref{tab:style-score}, where the poems generated with both general and style-biased topics are accounted for. It can be seen that overall, the style transfer may degrade fluency a little. This is understandable, as enforcing a particular style has to break the optimal generation with the baseline, which is assumed to be good at generating fluent poems. Nevertheless the sacrifice is not significant.

\begin{table}[htb]
	
	\centering
	\fontsize{6.5}{8}\selectfont
	\begin{tabular}{|l|c|c|c|c|c|c|}
		\hline
		Method               & Compliance &Fluency &Aesthetic   &Scenario\\
		&            &        &Innovation  &Consistence\\
		\hline
		$C_1$ (baseline)        &  4.10      &  3.01  &  2.53      & 2.94 \\
		$C_1$ + Pastoral Mem    &  4.07      &  3.00  &  3.07      & 3.17\\
		$C_1$ + Battlefield Mem &  3.82      &  2.63  &  2.60      & 2.95\\
		$C_1$ + Romantic Mem    &  4.00      &  2.78  &  2.59      & 3.00\\
		\hline
		$C_1$ + All Mem         &  3.95      &  2.80  &  2.74      & 3.05 \\
		\hline
	\end{tabular}
	\caption{\label{tab:style-score}Averaged scores for systems with
		or without style memory.}
	
\end{table}





\subsection{Examples}

Table~\ref{tab:ex1} to Table~\ref{tab:ex3} shows example poems generated by the system $C_1$, $C_1$+Mem  and $C_1$+Style Mem where the style in this case is set to be romantic. The three poems were generated with the same, very general, topic  (`(oneself)'). More examples are given in the supporting material.



\begin{table}[h]
\begin{center}
\begin{tabular}{|p{7cm}<{\centering}|}
\hline
\begin{CJK*}{UTF8}{gbsn}
自从此意无心物，
\end{CJK*}\\
Nothing in my heart,\\
\begin{CJK*}{UTF8}{gbsn}
一日东风不可怜。
\end{CJK*}\\
Spring wind is not a pity.\\
\begin{CJK*}{UTF8}{gbsn}
莫道人间何所在，
\end{CJK*}
\\
Don't ask where it is,\\
\begin{CJK*}{UTF8}{gbsn}
我今已有亦相传。
\end{CJK*}\\
I've noticed that and tell others.\\
\hline
\end{tabular}
\end{center}
\caption{\label{tab:ex1} Example poems generated by the $C_1$ system.}
\end{table}

\begin{table}[h]
\begin{center}
\begin{tabular}{|p{7cm}<{\centering}|}
\hline
\begin{CJK*}{UTF8}{gbsn}
一山自有无人语，
\end{CJK*}\\
Nobody speaking in the mountain,\\
\begin{CJK*}{UTF8}{gbsn}
不是青云入水边。
\end{CJK*}\\
Also no green cloud stepping into the river.\\
\begin{CJK*}{UTF8}{gbsn}
莫把春风吹落叶，
\end{CJK*}\\
Spring wind does not stir leaves,\\
\begin{CJK*}{UTF8}{gbsn}
花开绿树满江船。
\end{CJK*}\\
But flowers blooming in trees and flying to boats.\\
\hline
\end{tabular}
\end{center}
\caption{\label{ex-c1:cN} Example poems generated by the $C_1$+Mem system.}
\label{tab:ex2}
\end{table}

\begin{table}[h]
\begin{center}
\begin{tabular}{|p{7cm}<{\centering}|}
\hline
\begin{CJK*}{UTF8}{gbsn}
花香粉脸胭脂染，
\end{CJK*}\\
Beautiful face addressed by rouge,\\
\begin{CJK*}{UTF8}{gbsn}
帘影鸳鸯绿嫩妆。
\end{CJK*}\\
Mandarin duck outside the curtain.\\
\begin{CJK*}{UTF8}{gbsn}
翠袖红蕖春色冷，
\end{CJK*}\\
Green sleeves and red flowers in cold spring,  \\
\begin{CJK*}{UTF8}{gbsn}
柳梢褪叶暗烟芳。
\end{CJK*}\\
Willow leaves gone in fragrant mist.\\
\hline
\end{tabular}
\end{center}
\caption{\label{tab:ex3} Example poems generated by the $C_1$+Style Mem system where the style is romantic.}
\end{table}

\section{Conclusions}

In this paper, we proposed a memory mechanism to support innovative Chinese poem generation by neural models augmented with a memory. Experimental results demonstrated that memory can boost innovation from two opposite directions: either by encouraging creative generation for regularly-trained models, or by encouraging \mbox{rule-compliance} for overfitted models. Both strategies work well, although the former generated poetry that was preferred by experts in our experiments. Furthermore, we found that the memory can be used to modify the style of the generated poems in a flexible way. The experts we collaborated with feel that the present generation is comparable to today's experienced amateur poets. Future work involves investigating a better memory selection scheme. Other regularization methods (e.g., norm or drop out) are also interesting and may alleviate the over-fitting problem.

\section*{Acknowledgments}

This paper was supported by the National \mbox{Natural} Science Foundation of China (NSFC) under the project NO.61371136, NO.61633013, NO.61472428.

\newpage
\bibliography{acl2017}
\bibliographystyle{acl_natbib}

\end{document}